\documentclass{llncs}  

\usepackage{graphicx} 
\usepackage[T1]{fontenc}
\usepackage{amsmath}
\usepackage[english]{babel}
\usepackage{array,tabularx}
\usepackage{booktabs} 
\usepackage{makeidx}  
\usepackage{xparse}
\usepackage{moredefs}
\usepackage{lips}
\usepackage{epstopdf} 
\usepackage{color}
\usepackage{csquotes}
\usepackage{xcolor}
\usepackage{times}
\usepackage[hidelinks]{hyperref}
\usepackage[nolist,nohyperlinks]{acronym}
\usepackage{comment}
\usepackage{paralist}
\usepackage{cleveref}
\usepackage{wrapfig}
\usepackage{multirow}
\usepackage{microtype}
\usepackage{listings}
\usepackage[super]{nth}
\usepackage{natbib}
\usepackage{array}      
\usepackage{colortbl}
\usepackage{subcaption} 
\usepackage{siunitx}   

\DeclareCaptionLabelFormat{subfig-sf}{\sffamily(#2)}
\usepackage{float}
\floatstyle{plaintop}
\restylefloat{table}

\usepackage[style=base,labelfont=bf,labelsep=period,tableposition=top,font=small]{caption}
\makeatletter
\renewcommand\@biblabel[1]{#1.}
\makeatother
\addto\captionsenglish{}

\usepackage{fancyhdr}
\fancypagestyle{WI_footer}{\fancyhf{}\fancyfoot[L]{\color{black}\scriptsize 20th International Conference on Wirtschaftsinformatik\\September 2025, Münster, Germany}}

\crefformat{footnote}{#2\footnotemark[#1]#3}
\expandafter\def\expandafter\UrlBreaks\expandafter{\UrlBreaks
  \do\a\do\b\do\c\do\d\do\e\do\f\do\g\do\h\do\i\do\j%
  \do\k\do\l\do\m\do\n\do\o\do\p\do\q\do\r\do\s\do\t%
  \do\u\do\v\do\w\do\x\do\y\do\z\do\A\do\B\do\C\do\D%
  \do\E\do\F\do\G\do\H\do\I\do\J\do\K\do\L\do\M\do\N%
  \do\O\do\P\do\Q\do\R\do\S\do\T\do\U\do\V\do\W\do\X%
  \do\Y\do\Z}

\newcolumntype{L}[1]{>{\raggedright\arraybackslash}p{#1}}   
\newcolumntype{C}[1]{>{\centering\arraybackslash}p{#1}}     
\newcolumntype{R}[1]{>{\raggedleft\arraybackslash}p{#1}}    

\begin{document}
\frontmatter          

\mainmatter              

\title{Overcoming Algorithm Aversion with Transparency: Can Transparent Predictions Change User Behavior?}

\subtitle{Research Paper} 

\author{Lasse Bohlen\inst{1} \and
Sven Kruschel\inst{2} \and
Julian Rosenberger\inst{2} \and
Patrick Zschech\inst{1} \and
Mathias Kraus\inst{2}}

\institute{
Technische Universität Dresden, Dresden, Germany \\
\email{\{lasse.bohlen, patrick.zschech\}@tu-dresden.de}
\and
Universität Regensburg, Regensburg, Germany \\
\email{\{sven.kruschel, julian.rosenberger, mathias.kraus\}@ur.de}
}

\maketitle
\setcounter{footnote}{0}

\begin{abstract}

Previous work has shown that allowing users to adjust a machine learning (ML) model's predictions can reduce aversion to imperfect algorithmic decisions. However, these results were obtained in situations where users had no information about the model's reasoning. Thus, it remains unclear whether interpretable ML models could further reduce algorithm aversion or even render adjustability obsolete. In this paper, we conceptually replicate a well-known study that examines the effect of adjustable predictions on algorithm aversion and extend it by introducing an interpretable ML model that visually reveals its decision logic. Through a pre-registered user study with 280 participants, we investigate how transparency interacts with adjustability in reducing aversion to algorithmic decision-making. Our results replicate the adjustability effect, showing that allowing users to modify algorithmic predictions mitigates aversion. Transparency's impact appears smaller than expected and was not significant for our sample. Furthermore, the effects of transparency and adjustability appear to be more independent than expected. 
\end{abstract}


\thispagestyle{WI_footer}



\section{Introduction}
\label{sec:introduction}

As machine learning (ML) models are increasingly applied across domains, algorithmic decision-making is becoming more widespread \citep{janiesch_machine_2021, berger2021watch}. 
%
Despite those  models often demonstrating superior performance compared to human forecasters, individuals frequently exhibit algorithm aversion -- a reluctance to rely on algorithmic predictions \citep{dietvorst2015algorithm, dietvorst2018overcoming, jussupow_integrative_2024, mahmud2023drives}. While algorithm aversion is not strictly tied to performance differences, it is often observed even in cases where algorithms outperform humans.
 Researchers have identified multiple causes of this aversion \citep{jussupow_why_2020}, including desire for personal control \citep{dietvorst2018overcoming}, perceived inability to handle unique circumstances \citep{castelo2019task}, and sensitivity to algorithmic errors \citep{dietvorst2015algorithm, berger2021watch}.
A common concern is the black-box nature of many advanced ML models, which generate predictions without revealing their underlying reasoning. According to \cite{burton2020systematic}, achieving coherent decision-making requires aligning the human view with that of the algorithmic model, a process referred to as cognitive compatibility. This alignment is only achievable when the model is transparent enough for its reasoning to be understood. Without cognitive compatibility, algorithmic aids risk clashing with, rather than complementing, human intuition.

This observation has fueled the development of interpretable white-box models, which are fully transparent by design and thus aim to increase user acceptance \citep{rudin2019stop, kruschel2025challenging}. Interestingly, the empirical evidence on the effect of such models on users remains mixed. Particularly, \citet{poursabzi2021manipulating} found that study participants were not necessarily more likely to follow the recommendations of a simpler, interpretable model than a complex black-box model. These findings question the practical value of interpretability methods for overcoming algorithm aversion. An alternative approach, suggested by \citet{dietvorst2018overcoming}, involves giving users limited control over algorithmic model outputs. In their experiment, participants are allowed to make small adjustments to a model's prediction. This possibility of adjustments significantly increased the willingness to use the algorithmic aid and improved the overall performance of the prediction. However, their study was conducted in a context where participants received no information about the model's decision-making process. This leaves open the question of whether transparency of interpretable models might serve as a substitute for direct control -- or whether transparency and adjustability might work synergistically to overcome algorithm aversion.


In our work, we deepen the understanding of the relationship between algorithm aversion, interpretability, and user control. Through a pre-registered user study with 280 participants, we make the following two primary contributions. First, we test the robustness of \citet{dietvorst2018overcoming}'s findings through a conceptual replication of their study using a different prediction task. Second, we extend their paradigm by examining how algorithm transparency through visual explanations of the model's decision logic influences users' willingness to rely on algorithmic predictions, with and without the ability to adjust those predictions.

Our results show that providing participants with the ability to make adjustments to a model's predictions significantly reduces algorithm aversion, closely replicating the findings of \citet{dietvorst2018overcoming}. In contrast, transparency alone, implemented in the form of visual explanations of an interpretable model's feature contributions, shows only a modest and statistically insignificant increase in participants' willingness to choose the model. As a result, while participants in our transparency (i.e., \emph{white-box}) condition showed slightly lower error rates, transparency alone did not reliably increase model usage. These findings support the idea that providing users with a tangible sense of control may be more important in overcoming aversion to algorithms than simply revealing the inner workings of the model.

The remainder of this paper is structured as follows. Section \ref{sec:background} provides an overview over related literature on algorithm aversion, interpretable ML, and approaches to overcoming user resistance. Section \ref{sec:method} details our experimental design and analysis. Section \ref{sec:results} presents our findings, and Section \ref{sec:discussion} discusses their implications for theory and practice.

\section{Research Background}
\label{sec:background}
\subsection{Algorithm Aversion}
\label{sec:aversion}

Algorithm aversion describes people's reluctance to use algorithmic predictions \citep[e.g., ][]{esteva2017dermatologist, brown2019superhuman}. Naturally, this phenomenon poses significant challenges in the adoption of algorithmic decision support systems and reasons for this are manifold \citep{jussupow_why_2020}. In general, people often express less satisfaction with decisions when they learn that they were made by algorithms rather than humans \citep{longoni2019resistance}. This might be due to users' desire for perfect predictions and their low tolerance for an algorithm's mistakes \citep{dietvorst2015algorithm, dietvorst2018overcoming, wanner2022effect}. Additionally, subjective or emotion-based tasks contribute to algorithm aversion \citep{castelo2019task, germann2023algorithm}, as people perceive that algorithms fail to consider individual circumstances and characteristics. This aversion is further amplified by ethical concerns, particularly when automated advice influences high-risk decisions \citep{longoni2019resistance}. Surprisingly, also in cases where algorithmic decisions lead to better outcomes they are often perceived as black-box and thus difficult to understand, ultimately leading to algorithm aversion \citep{cadario2021understanding, filiz2021reducing}. 

As a remedy, researchers have developed several strategies to mitigate algorithm aversion. Key approaches include providing choice over the training algorithm \citep{cheng2023overcoming}, framing algorithms as considering individual characteristics \citep{longoni2019resistance} or give users control over algorithmic outputs \citep{dietvorst2018overcoming}. The ability for users to adjust predictions directly addresses the desire for personal control, offering a tangible sense of control and potentially fostering psychological ownership over the final decision \citep{pierce2001theorya}. Furthermore, this adjustability empowers users with the ability for perceived error correction of the model's suggestions, thereby enhancing their overall sense of control in the decision-making process \citep{dietvorst2018overcoming}. Another promising approach is to increase model transparency. \citet{mahmud2022influences} found that the inherent characteristics of prediction models themselves can significantly influence algorithm aversion. Therefore, in this work, we investigate the effect of model transparency on algorithm aversion. Expanding this topic, the following section examines interpretable models as a promising way to empower users to understand an algorithm's decision-making.


\subsection{Interpretable Models}

In order to obtain transparent ML models, two different approaches are applicable: post-hoc explanation methods or white-box models  \citep{rudin2019stop}. Post-hoc explanation methods like LIME \citep{ribeiro2016should} or SHAP \citep{lundberg2017unified} can explain complex models after they were fit to a specific task. In general, post-hoc explanations provide only approximate insights, adding uncertainty and potentially leaving users’ concerns unaddressed \citep{rudin2019stop}. In contrast, interpretable models are designed to be transparent in their operation by constraining their complexity. Thus, their decision-making process can be directly examined and understood in detail without additional tools. Linear regression is one of the most basic examples for interpretable models, where each feature's contribution to the prediction is directly quantifiable through its coefficient. Generalized additive models (GAMs) extend linear models by allowing non-linear relationships between predictors and the response variable, while retaining an additive structure \citep{kraus2024interpretable}. When creating predictions with GAMs the effects of different features remain separate and can be visualized independently, allowing users to understand exactly how each variable influences the prediction \citep{kruschel2025challenging}. By using an algorithm that can visually communicate its decision logic, we test whether transparency can actually help to reduce algorithm aversion.

\subsection{Overcoming Aversion with Interpretable Models}

Research suggests that more interpretable models can reduce algorithm aversion because users who understand how predictions are made feel less uncertainty about the algorithm’s reasoning, making them more likely to accept and rely on the model \citep{miller2019explanation, binns2018s, aslan2024mitigating}. Unlike human decision-makers, who can explain their reasoning, algorithmic decision aids often provide little to no justification, making users hesitant to rely on them \citep{kayande2009incorporating}. Studies have shown that transparent models can enhance trust and acceptance in algorithmic decision-making \citep{leichtmann2023effects, wanner2022effect}. Ideally, transparency not only builds trust but also enables users to more accurately assess when the algorithm is likely to be correct and when its advice should be questioned or modified \citep{zerilli2022transparency}. \citet{yeomans2019making} demonstrated that algorithm aversion decreases when users can comprehend the model's reasoning. Opening the "black-box" can improve acceptance because transparency helps users understand algorithmic logic \citep{litterscheidt2020financial, mahmud2022influences}, potentially creating cognitive compatibility between human judgment and algorithmic decision logic \citep{burton2020systematic}. However, transparency has limitations. Work by \citet{poursabzi2021manipulating} suggests, that interpretability alone does not necessarily mitigate algorithm aversion. If explanations are too complex, they may create only an illusion of understanding rather than genuine insight.

Given the distinct mechanisms through which adjustability enhances perceived control, and transparency aims to improve understanding and trust, their potential interplay in influencing algorithm aversion is critical to understand. For instance, high transparency, by reducing uncertainty and improving understanding, might diminish the perceived need for direct control offered by adjustability (a substitutive effect). Alternatively, enhanced understanding from transparency could empower users to make more effective adjustments, leading to a stronger combined impact (a synergistic effect).

Building on this background, our study proceeds with several key expectations. First, in line with the original work by \citet{dietvorst2018overcoming}, we anticipate that providing users the ability to adjust algorithmic predictions reduces algorithm aversion (manifested as increased model usage) and also lead to improved task performance. Secondly, we hypothesize that introducing transparency into the model's decision logic encourages users to rely more on the algorithm, enhancing their task performance.

\section{Research Approach}
\label{sec:method}

\subsection{Prediction Task}
We study algorithm aversion in a context that participants can intuitively grasp, selecting the UCI Machine Learning Bike Sharing dataset.\footnote{\href{https://archive.ics.uci.edu/ml/datasets/Bike+Sharing+Dataset}{https://archive.ics.uci.edu/ml/datasets/Bike+Sharing+Dataset}} This dataset tracks hourly bike rentals based on weather and calendar-related features. In the experiment, participants view these features and predict the number of rentals for a given day and time. From the full dataset, we choose six features that include both continuous (temperature, windspeed, and humidity) as well as categorical (time of day, type of day, and weather situation) variables. We focus on these features because they are easily interpretable and most people intuitively understand how weather and time variables might affect bike rentals.

\subsection{Implementation of the Interpretable Model}
\begin{figure}[h!]
\includegraphics[width=\textwidth]{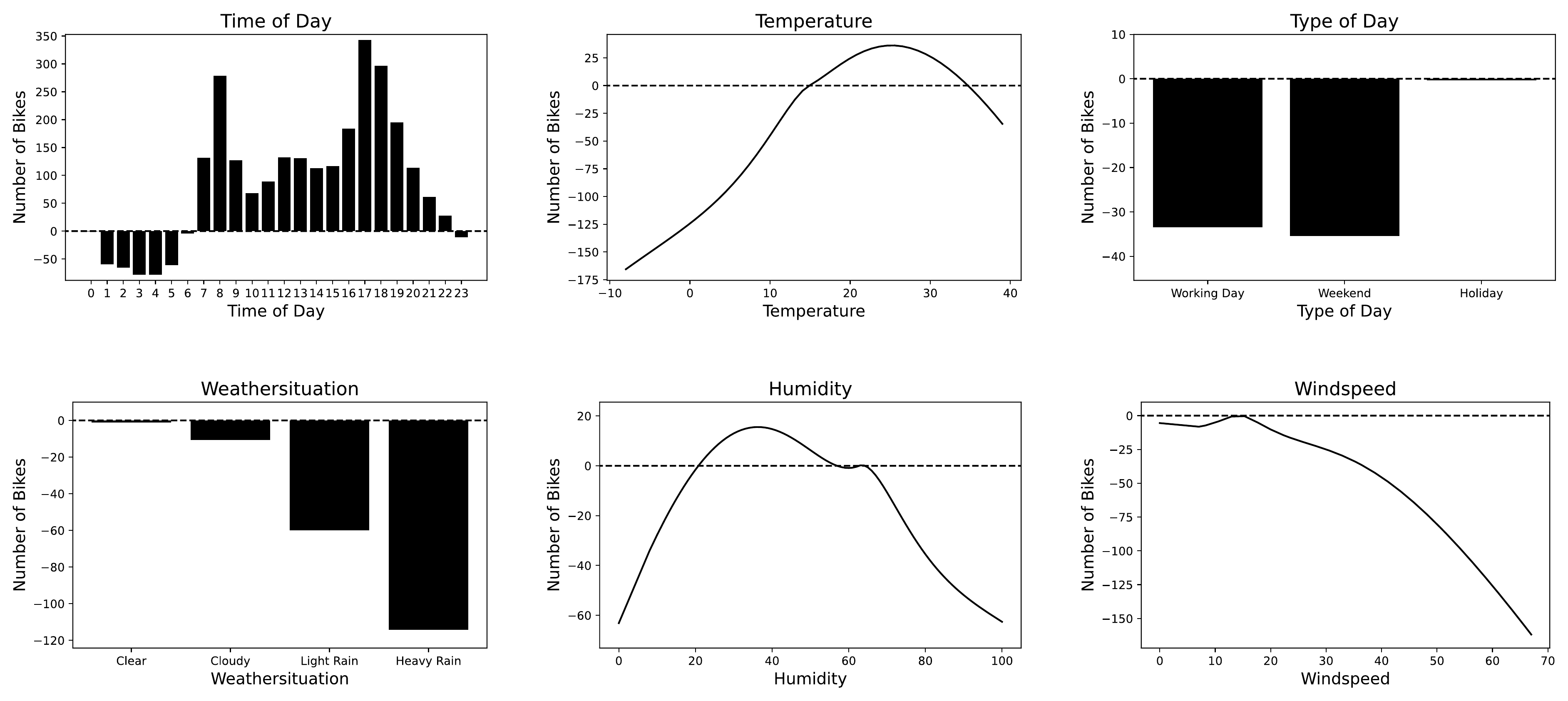}
\caption{Visualization showing the learned algorithm's decision logic between the individual features and the target (bike demand) of the GAM.}
\label{fig:iGANN_overview}
\end{figure}

In \citet{dietvorst2018overcoming}, algorithmic predictions came from a simple linear regression model, but the study focused on algorithm aversion under black-box conditions rather than revealing the model’s logic to participants. While linear regression provides interpretable coefficients, it may oversimplify complex relationships. Since our analysis requires making the model’s decision-making process visually interpretable, we address these limitations by using a GAM \citep{kruschel2025challenging}. A GAM expresses each feature’s contribution with its own potentially non-linear function $f_i$,
\begin{equation}
    \hat{y} = \beta_0 + f_1(x_1) + f_2(x_2) + \cdots + f_n(x_n).
\end{equation}
This structure allows more flexible modeling of relationships while maintaining interpretability. Users can examine each \(f_i\) independently to see how each feature influences the prediction.
We implement our GAM based on IGANN \citep{kraus2024interpretable}. We train IGANN on the bike-sharing dataset, focusing on the six selected features. The final model results in a mean absolute error of around 80, which we communicate to participants so that they are aware of the model's imperfection. Notably, this error is similar in magnitude to the error of the linear model in the original study. The feature plots of the trained model are shown in Figure~\ref{fig:iGANN_overview}. The structure of the GAM makes it easy for participants to see how each variable contributes to the predicted number of bike rentals, a relationship that is generally intuitive.

\subsection{User Study}
The goal of our user study is to examine how transparency from interpretable models affects algorithm aversion. Building on \citet{dietvorst2018overcoming}, participants complete a prediction task under conditions that systematically vary their control over the model’s outputs (i.e., whether they can adjust them) and their insight into the model’s logic (i.e., with or without transparent model structure). This setup tests whether the model's transparency amplifies or diminishes the known effect that minimal user control reduces aversion to algorithms. This study was pre-registered\footnote{\href{https://doi.org/10.17605/OSF.IO/RX2TS}{https://doi.org/10.17605/OSF.IO/RX2TS}} and received ethical approval\footnote{https://gfew.de/ethik/8wkDVk4x} before it was carried out.
The remainder of this section covers three key elements. First, the treatment specification explains how participants can adjust the predictions or observe the algorithm’s reasoning. Second, the study procedure is outlined and finally, the analysis methods are detailed.

\subsubsection{Treatments.}
We employ a 3 × 2 between-subjects factorial design, through a conceptual replication and extension of the \emph{overcoming algorithm aversion} study of \citet{dietvorst2018overcoming} using the bike rental prediction task. Participants view contextual features and predict hourly rental demands to earn a performance-based bonus. The first treatment factor, adjustability, has three levels. In the \emph{can't-change} condition, participants must accept to rely on the model's prediction or disregard it entirely. In \emph{adjust-by-50}, they can shift the model's prediction by up to 50 bikes or ignore it completely. In \emph{use-freely}, they can change the model's suggestion by any amount. The second treatment factor, transparency, has two levels. In the \emph{white-box} condition, participants see visualizations of the algorithm’s decision logic via GAM feature plots. In the \emph{black-box} condition, they get no insight into the model's internal decision logic.

\subsubsection{Study Procedure.}
The study is conducted via an online survey. Participants are introduced to the prediction task, receiving detailed descriptions of the features and learning that demand ranges from 0 to 1,000 bikes, with an average demand of 190 bikes. Next, participants are randomly assigned to one of six conditions in the 3~×~2 experimental design. They are then introduced to the ML model designed to predict bike rental demand. They learn that the model is trained on real data, uses the same features available to them, and has an average error of 80 bikes. Participants in the newly introduced \textit{white-box} condition also receive GAM feature plots visualizing the model's internal structure to help them understand its decision logic (cf. Figure \ref{fig:iGANN_overview}). All participants are then informed of the incentive structure: they can earn a bonus of up to \textsterling 5 based on the accuracy of their predictions, with the bonus decreasing for larger errors.
To ensure comprehension, participants are asked to type a sentence summarizing their adjustability condition and the incentive structure. 
Next, participants in the \emph{can't-change} and \emph{adjust-by-50} conditions must make a one-time binary choice between using the model's predictions or not. They make this choice once and right before they begin making their 20 predictions.
Following \citet{dietvorst2018overcoming} procedure, no binary choice is presented in the \emph{use-freely} condition, since participants can fully adjust the model’s predictions, making an explicit decision meaningless.

\begin{figure}[h]
    \centering
    \includegraphics[width=1\linewidth]{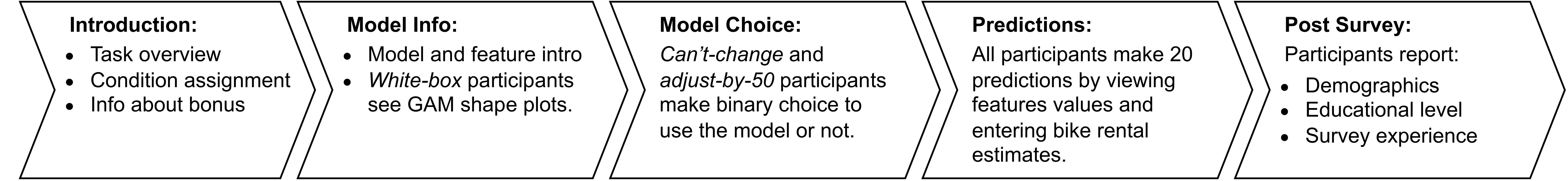}
    \caption{Schematic diagram of the survey procedure}
    \label{fig:survey_flow}
\end{figure}

Following this decision, all participants make 20 predictions. In the \emph{use-freely} condition, participants can modify the model’s prediction as they wish before submitting their final estimate. In the \emph{adjust-by-50} condition, those who choose to rely on the model can adjust its prediction by up to 50 bikes. In the \emph{can’t-change} condition, participants who rely on the model use its prediction directly to calculate their bonus, with no adjustments allowed. In both the \emph{adjust-by-50} and \emph{can’t-change} conditions, participants who do not rely on the model make predictions entirely on their own, without seeing the model’s output.
Figure \ref{fig:survey_flow} shows a schematic diagram of the study. The complete online study for all treatments can also be viewed in our online repository.\footnote{\href{https://doi.org/10.17605/OSF.IO/RMNFH}{https://doi.org/10.17605/OSF.IO/RMNFH}}

\subsubsection{Participants.}
In total, we recruited 300 participants through Prolific. Key inclusion criteria set on the Prolific platform included an approval rating of 99-100\%, English as a first language, and balanced quotas for male and female participants. Participants were recruited from all available countries on the platform. The study lasted an average of 20 minutes and 39 seconds, with participants being paid a base rate and additional bonus payments, resulting in an average of £13 per hour. To ensure data quality, we included two attention checks, resulting in the exclusion of 20 participants. The final sample comprised 280 participants (50.8\% female, 47.5\% male, one non-binary, and 2 undisclosed). Randomization checks confirmed balanced demographic distribution across treatment groups. Fisher’s exact test showed no significant gender differences ($p = 0.498$), and ANOVA revealed no significant age disparities (adjustability: $F(2,\,280) = 0.721, p = 0.487$; transparency: $F(1,\,280) = 1.152, p = 0.284$), ensuring comparability across groups. The age and gender distribution for the individual treatment groups is shown in Table \ref{tab:demographics}.
\begin{table}[h]
\scriptsize
\centering
\renewcommand{\arraystretch}{1.0}
\setlength{\tabcolsep}{6pt} 

\begin{tabular}{c c c c c c c}
\toprule
 {adjustability} & {transparency} & {Female} & {Male} & {Other} & {Total} & {Age (Mean ± SD)} \\ 
\midrule
 \emph{can't-change}   & \emph{white-box}  & 27  & 25  & 0 & 52 & 39.2 ± 12.3  \\
 \emph{can't-change}   & \emph{black-box}  & 19  & 21  & 0 & 40 & 35.3 ± 11.4  \\
 \addlinespace[5pt]
 \emph{adjust-by-50}   & \emph{white-box}  & 22  & 23  & 0 & 45 & 35.9 ± 10.3  \\
 \emph{adjust-by-50}   & \emph{black-box}  & 18  & 25  & 0 & 43 & 37.1 ± 13.7  \\
 \addlinespace[5pt]
 \emph{use-freely}    & \emph{white-box}  & 24  & 19  & 1 & 44 & 39.8 ± 14.1  \\
 \emph{use-freely}    & \emph{black-box}  & 27  & 27  & 2 & 56 & 37.8 ± 13.7  \\
\midrule
 {Overall}       & - & {137} & {140} & {3} & {280} &{37.5 ± 12.6}  \\
\bottomrule
\end{tabular}
\caption{Descriptive statistics for gender and age across adjustability and transparency treatments}
\label{tab:demographics}
\end{table}

\subsubsection{Analysis.}
We analyze two primary outcomes to examine how transparency and adjustability affect users' behavior in algorithmic predictions: (i) model choice and (ii) task performance measured as mean absolute error from actual rental demand. We use mean absolute error rather than model deviation because the ML model typically outperforms participants and most deviations increase error rather than improve predictions. However, 68 out of 280 participants managed to outperform the model -- remarkably, 67 of them had access to its predictions. This suggests that while the algorithm generally provides strong predictive performance, users can improve upon it under certain conditions. Specifically, access to the model’s predictions appears to enable users to identify and correct its shortcomings, leading to better outcomes. In this context, using model deviation as a measure might misinterpret these beneficial adjustments as algorithm aversion, even though they reflected productive engagement with the model.
However, additional analyses using the deviation from the model and bonus earned yield similar results and are included in our online appendix.\footnote{\href{https://doi.org/10.17605/OSF.IO/RMNFH}{https://doi.org/10.17605/OSF.IO/RMNFH}}

For the model choice analysis, we used chi-squared tests to assess whether adjustability (\emph{can't-change}, \emph{adjust-by-50}) significantly affected users' decisions to rely on the algorithm. We also analyzed how transparency (\emph{white-box} vs. \emph{black-box}) affected this choice. To analyze the task performance, we performed a two-way ANOVA with transparency and adjustability as factors. This allowed us to identify the potential main effects of each treatment as well as their potential interaction. Before running the ANOVA, we check key assumptions, including normality of residuals and homogeneity of variance. For post-hoc comparisons, we used pairwise t-tests with Bonferroni adjustments to control for multiple comparisons.

\section{Results}
\label{sec:results}

\subsection{Model Choice}
\label{sec:results:modelchoice}
\begin{figure}[h]
    \centering
    \begin{subfigure}[b]{0.43\textwidth}
        \centering
        \includegraphics[width=\textwidth]{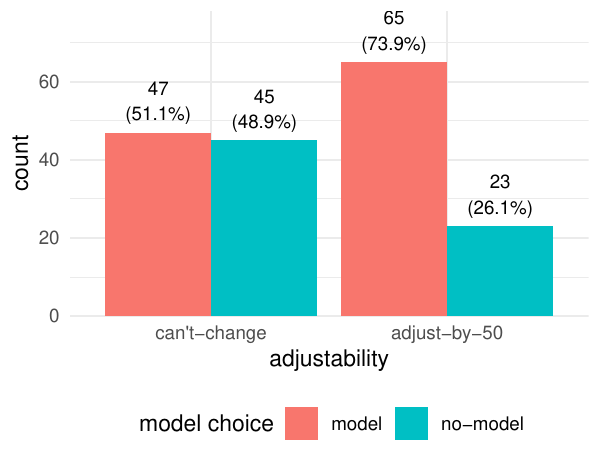}
        \caption{\sffamily Participants' model choice by adjustability. Participants allowed to make adjustments chose the model more (73.9\%) than those who couldn't (51.1\%), $\chi^2(1, N = 180) = 8.98, p = .003$.}
        \label{fig:ModelChoice(a)}
    \end{subfigure}
    \hfill 
    \begin{subfigure}[b]{0.43\textwidth}
        \centering
        \includegraphics[width=\textwidth]{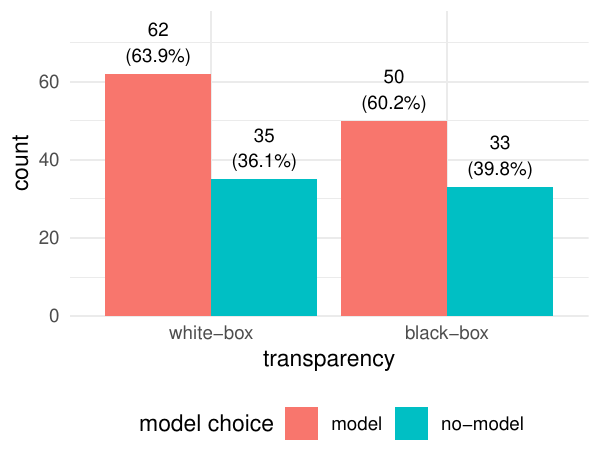}
        \caption{\sffamily Participants' model choice by transparency. Only minimal difference between \emph{white-box} (63.9\%) and \emph{black-box} algorithm recognizable (61.0\%), $\chi^2(1, N = 180) = 0.06, p = .724$.}
        \label{fig:ModelChoice(b)}
    \end{subfigure}
    \caption{Influence of adjustability (a) and transparency (b) on model choice.}
    \label{fig:ModelChoice}
\end{figure}

\cite{dietvorst2018overcoming} demonstrate that giving users the option to adjust an algorithm's predictions significantly increases their willingness to rely on that algorithm. Our results replicate this finding clearly. As shown in Figure \ref{fig:ModelChoice}, Participants who had the option to adjust predictions (\textit{adjust-by-50}) chose to rely on the model more frequently (65 participants, 73.9\%) compared to those in the \textit{can't-change} condition (47 participants, 51.1\%). A chi-square test confirmed this difference was statistically significant ($\chi^2(1,\, N = 180) = 8.98, p = .003$). Thus, our findings support the notion that providing users with limited control over algorithmic predictions substantially reduces algorithm aversion. In contrast, transparency alone showed a smaller and non-significant effect on participants' choice. Specifically, in the \emph{white-box} condition, 62 participants (63.9\%) chose to rely on the model. In the \emph{black-box} condition, 50 participants (60.2\%) chose the model. However, this difference did not reach statistical significance ($\chi^2(1,\,N = 180) = 0.12, p = .724$). Thus, transparency alone was insufficient to meaningfully reduce algorithm aversion without providing users with the ability to adjust predictions. This result differs from our expectation: We assumed that disclosing how the algorithm works would significantly increase adoption, even or especially if users could not adjust its predictions.

\subsection{Task Performance}
\begin{table}[h]
\scriptsize
\centering
\caption{Descriptive statistics for mean task error across adjustability and transparency conditions}
\label{tab:results mean error}
\renewcommand{\arraystretch}{1.1}
\setlength{\tabcolsep}{6pt} 
\sisetup{
       separate-uncertainty,
       table-number-alignment=center,
       print-zero-integer=true       
    }
\begin{tabular}{l l 
    S[table-format=3.1(3)] c
    S[table-format=3.1(3)] c
    S[table-format=3.1(3)] c}
\toprule
 & & \multicolumn{4}{c}{transparency} & & \\
\multirow{2}{*}{} & & \multicolumn{2}{c}{\emph{white-box}} & \multicolumn{2}{c}{\emph{black-box}} & \multicolumn{2}{c}{{Overall}} \\
\cmidrule(lr){3-4} \cmidrule(lr){5-6} \cmidrule(lr){7-8}
 & & {Mean ± SD\hspace{5pt}} & {N} & {Mean ± SD\hspace{5pt}} & {N} & {Mean ± SD\hspace{5pt}} & {N} \\
\midrule \addlinespace[5pt]
\multirow{3}{*}{\rotatebox{90}{adjustability}}  
    & \emph{can't-change}  & \num{121.6(51.9)}  & 52  & \num{137.0(61.9)}  & 40  & \num{128.3(56.7)}  & 92 \\ \addlinespace[5pt]
    & \emph{adjust-by-50}  & \num{105.8(42.9)}  & 45  & \num{113.4(64.9)}  & 43  & \num{109.5(54.6)}  & 88 \\ \addlinespace[5pt]
    & \emph{use-freely}    & 96.6(20.3)  & 44  & \num{102.7(29.3)}  & 56  & \num{100.0(25.8)}  & 100 \\ \addlinespace[5pt]
\midrule
& Overall & \num{108.8(42.4)}  & 141 & \num{115.9(53.9)}  & 139 & \num{112.3(48.2)}  & 280 \\
\bottomrule
\end{tabular}
\end{table}

\noindent Table \ref{tab:results mean error} shows the mean error for each treatment and as average across all treatment groups. We analyze task performance in terms of mean absolute prediction error, using a two-way ANOVA (cf. Table~\ref{tab:anova}), with adjustability and transparency as independent factors. Consistent with previous research, the results of the two-way ANOVA indicate a significant main effect from the adjustability treatment~($F(2,\,N=280) = 9.5$, \mbox{$p < .001$}). However, against our expectations, transparency alone did not significantly affect task performance ($F(1,\, N=280) = 2.89$, $p = .09$) and the effect size suggests only a small impact in our sample. A possible interaction between both treatment dimensions is not significant and negligible, ($F(2,\,280) = 0.22$, $p = .807$), indicating the effects of adjustability did not depend on transparency or the other way around. Therefore, we analyze both treatments separately below.

\begin{table}[h]
\scriptsize
\centering
\caption{Results of a two-way ANOVA reporting the effect of adjustability and transparency on participants' mean task error.}
\label{tab:anova}
\renewcommand{\arraystretch}{1.2}
\setlength{\tabcolsep}{10pt} 
\begin{tabular}{
  l
  S[table-format=1.0]
  S[table-format=2.3]
  S[table-format=<1.3]
  S[table-format=1.3]
}
\toprule
{Treatment} & {df} & {$F$} & {$p$-value} & {Effect size $\eta_p^2$} \\ 
\midrule
{adjustability}               & 2 & 9.500 & {$<$}0.001 & 0.065 \\
{transparency}                & 1 & 2.893 & 0.090    & 0.010 \\
{adjustability $\times$ transparency} & 2 & 0.255 & 0.775    & 0.002 \\ 
\bottomrule
\end{tabular}
\end{table}

\subsubsection{Effect of Adjustability.}
As the ANOVA shows that there are significant differences between the groups' mean task error in terms of adjustability ($\eta^2 = 0.065$, \mbox{$p < .001$}), we perform a post-hoc test in the form of a paired t-test with Bonferroni correction to further examine this result (cf.~Table~\ref{tab:posthoc}).
It reveals that participants in the \textit{can't-change} condition have significantly higher errors (128.3 $\pm$ 56.7) than those in the \textit{adjust-by-50} (109.5 $\pm$ 54.6, $p = 0.008$) and \textit{use-freely} conditions (100.0 $\pm$ 25.8, $p <0.001$) while the difference between \textit{use-freely} and \textit{adjust-by-50} is not significant ($p = 0.170$) (cf. Figure \ref{fig:MErrorAdjustability}). These findings align with \citet{dietvorst2018overcoming}, as the mere ability to adjust a prediction increases the likelihood of choosing the model, ultimately leading to better performance. However, since participants generally struggle to outperform the predictions from the model, the intensity of possible adjustments has little impact.

\begin{table}[h]
\scriptsize
\centering
\caption{Results of post-hoc pairwise t-test comparisons for mean task error between different adjustability treatments}
\label{tab:posthoc}
\renewcommand{\arraystretch}{1.2}
\setlength{\tabcolsep}{10pt}
\begin{tabular}{
  l
  l
  S[table-format=2]
  S[table-format=2]
  S[table-format=<1.3]
  S[table-format=<1.6]
}
\toprule
{Group 1} & {Group 2} & {$n_1$} & {$n_2$} & {$p$-value} & {Adjusted $p$-value} \\ 
\midrule
\emph{can't-change} & \emph{adjust-by-50} & 92 & 88 & 0.008 & 0.024\,{(*)} \\
\emph{can't-change} & \emph{use-freely}  & 92 & 100 & {$<$}0.001 & {$<$}0.001\,{(***)} \\
\emph{adjust-by-50} & \emph{use-freely}  & 88 & 100 & 0.170 & 0.511\,{(ns)} \\ 
\bottomrule
\end{tabular}
\end{table}

\subsubsection{Effect of Transparency.}
\begin{figure}[h!]
    \centering
    \begin{subfigure}[b]{0.45\textwidth}
        \centering
        \includegraphics[width=\textwidth]{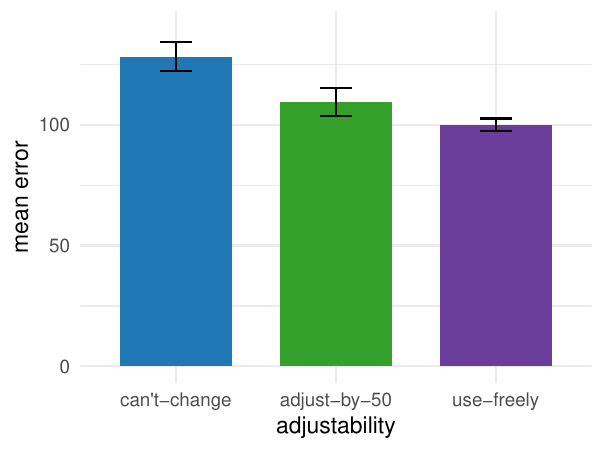}
        \caption{\sffamily Mean error across adjustability. Errors are significantly higher in the can't-change condition compared to adjust-by-50 ($p = 0.008$) and use-freely ($p < 0.001$).}
        \label{fig:MErrorAdjustability}
    \end{subfigure}
    \hfill 
    \begin{subfigure}[b]{0.45\textwidth}
        \centering
        \includegraphics[width=\textwidth]{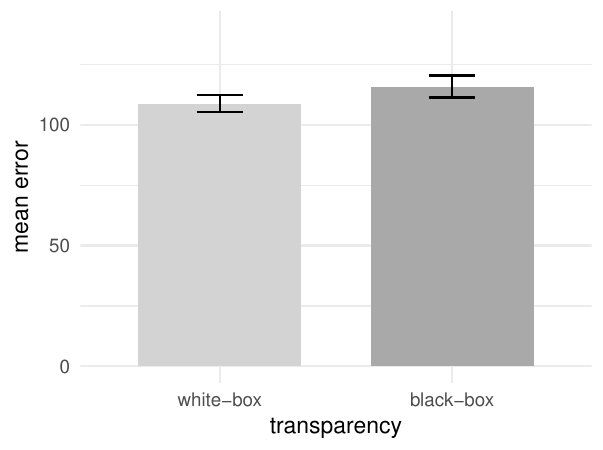}
        \caption{\sffamily Mean error across transparency conditions. While white-box participants show average lower errors, the difference is not statistically significant ($p = 0.110$).}
        \label{fig:MErrorTransparency}
    \end{subfigure}
    \begin{subfigure}[b]{0.90\textwidth}
        \centering
        \includegraphics[width=\textwidth]{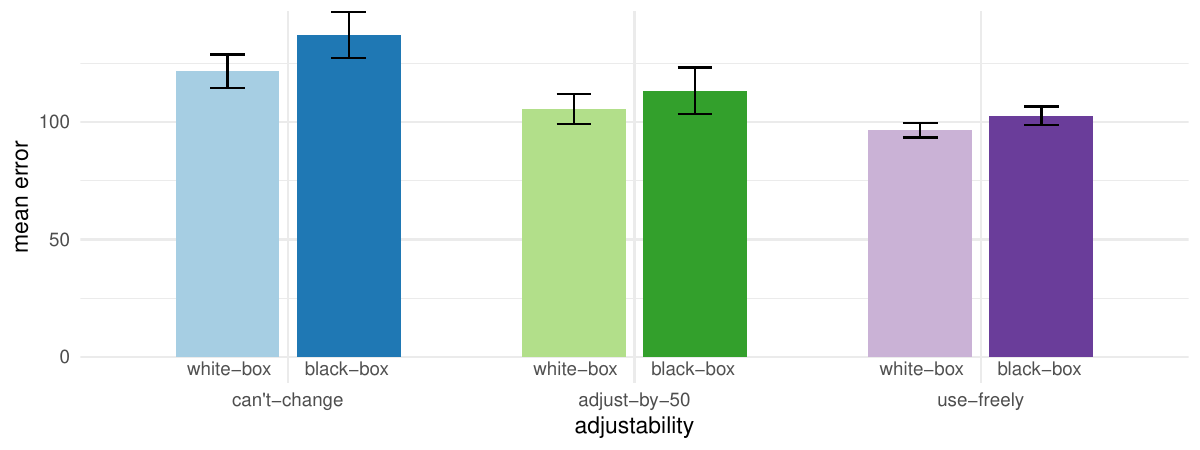}
        \caption{\sffamily Mean error for both transparency and adjustability treatments. The trend suggests slightly lower errors in the white-box condition across all levels of adjustability.}
        \label{fig:MErrorTransparencyAdjustability}
    \end{subfigure}
    \caption{Mean error across experimental conditions. (a) By adjustability treatment, (b) by transparency treatment, and (c) across the combined adjustability and transparency treatments}
    \label{fig:MError}
    \vspace{-10pt}
    
\end{figure}

Although the ANOVA finds no significant differences in mean prediction accuracy between \emph{black-box} and \emph{white-box} participants (cf.~Table~\ref{tab:anova}), descriptive analysis reveals a consistent trend: participants in the \emph{white-box} condition exhibit lower average task errors, both overall and within each adjustability subgroup (see Figure \ref{fig:MErrorTransparency} and \ref{fig:MErrorTransparencyAdjustability}). One possible explanation is that participants who see the model’s decision logic do not necessarily choose it more often (cf. Section~\ref{sec:results:modelchoice}) but may rely on it more when making adjustments, leading to improved accuracy. Additionally, by interpreting the visual feature plots of the transparent model, they may develop a better understanding of the relationships between predictors and bike demands, enabling them to make more informed predictions even without directly seeing the model’s output during the prediction task. To further examine this trend, we conducted a directed post-hoc t-test. The results indicate that mean error in the \emph{white-box} condition ($M = 108.75$) was lower than in the \emph{black-box} condition ($M = 115.90$). However, this difference did not reach statistical significance ($t(261.62) = -1.23, p = 0.110$). 

\section{Discussion}
\label{sec:discussion}
In this work, we explore whether revealing a prediction model’s decision logic reduces aversion and how transparency and adjustability interact to influence user behavior. Our findings confirm that adjustability significantly reduces algorithm aversion and improves task performance \citep{dietvorst2018overcoming}. In contrast, transparency alone has little impact on task performance and none on model choice. Although, participants in the \emph{white-box} condition show lower task errors, this trend is not statistically significant. A power analysis showed that to detect such small effects of transparency (partial $\eta^2 \approx 0.01$, Cohen's $f \approx 0.10$), 1,289 participants would be needed for 95\% power in the two-way ANOVA. Further, for a significant post hoc t-test, approximately 2,000 participants would be required (Cohen's $d \approx 0.15$, $\alpha = 0.05$, 95\% power). A possible explanation for the weak effect of transparency is that simply revealing the model’s decision logic once does not guarantee users will engage with or understand it. Moreover, the global nature of the feature plots, while offering general insights into the model's logic, may not have adequately supported task-specific sense-making required by participants when deciding on or adjusting individual predictions. Participants in the \emph{white-box} condition may see the visualizations but not actively process them. Unlike adjustability, which directly alters outcomes, transparency requires cognitive effort, which users may disregard if they see no immediate benefit. Notably, participants view the model’s feature plots before knowing they will choose whether to use the algorithm. Even with transparency, people may favor their intuition over algorithmic predictions. Research shows that individuals trust human judgment more, especially in tasks where they feel confident \citep{yeomans2019making}. If participants believe their judgment matches the algorithm’s, interpretable visualizations alone may not shift their reluctance toward the algorithm. This suggests that algorithm aversion may be more effectively mitigated by interventions that enhance user adjustability rather than by simply increasing transparency.


Our findings have implications for the design of algorithmic ML-based decision support systems. First, the results indicate that providing users with some degree of control over algorithmic outputs is more effective than transparency alone. Systems designed to assist decision-making should integrate interactive features that allow users to adjust, refine, or personalize algorithmic predictions \citep{sele2024putting}. Second, the limited impact of transparency in our study suggests that the way explanations are presented matters. Rather than passively displaying model visualizations, systems may benefit from interactive explanations that encourage users to actively explore how the model makes predictions. Research in interpretable ML increasingly suggests that engagement-driven transparency, where users can manipulate input variables and observe changes in predictions, is more effective than static explanations \citep{wang2022interpretability}.

Our study has limitations that should be acknowledged. First, we do not directly measure participants’ engagement with the transparency treatment. While they see visual feature plots of the model’s decision logic, we do not assess whether they actively engage with this information. Future research should incorporate measures such as self-reported engagement levels to better measure this engagement. Second, our experiment takes place in a single session. Longitudinal studies could explore whether repeated exposure to interpretable models influences reliance over time. Additionally, our study focuses on a relatively simple prediction task. Future research should therefore examine whether transparency has a stronger effect in complex domains where users inherently struggle to rely on algorithmic reasoning, such as in medical decision-making. Finally, future work could use the Judge-Advisor System paradigm to more precisely quantify how individuals incorporate algorithmic advice by measuring the extent to which participants adjust their initial predictions after viewing the model’s output \citep{bonaccio2006advice}.



\bibliographystyle{agsm}
\bibliography{literature}

\end{document}